\documentclass[runningheads]{llncs}
\usepackage[T1]{fontenc}
\usepackage{graphicx,xcolor}
\usepackage{algorithm,algpseudocode}
\begin{document}
\title{A Boosting Approach to Constructing an Ensemble Stack\thanks{Supported by 2Keys Corporation - An Interac Company.}}
%
%
\author{Zhilei Zhou\inst{1} \and
Ziyu Qiu\inst{1} \and
Brad Niblett\inst{2} \and
Andrew Johnston\inst{2}\and 
Jeffrey Schwartzentruber\inst{1} \and
Nur Zincir-Heywood\inst{1}\orcidID{0000-0003-2796-7265} \and
Malcolm I. Heywood\inst{1}\orcidID{0000-0002-1521-0671}
}
\authorrunning{Zhou et al.}
%
\institute{Faculty of Computer Science, Dalhousie University, Nova Scotia, Canada\\
\email{\{ZhileiZhou,amousqiu,zincir,mheywood\}@dal.ca,\\jeffrey.schwartzentruber@gmail.com} \and
2Keys Corporation - An Interac Company, Ottawa, Canada
\email{\{bniblett,ajohnston\}@2keys.ca}\\
\url{http://www.2keys.ca}}
\maketitle              
\begin{abstract}
An approach to evolutionary ensemble learning for classification is proposed in which boosting is used to construct a stack of programs. Each application of boosting identifies a single champion and a residual dataset, i.e. the training records that thus far were not correctly classified. The next program is only trained against the residual, with the process iterating until some maximum ensemble size or no further residual remains. Training against a residual dataset actively reduces the cost of training. Deploying the ensemble as a stack also means that only one classifier might be necessary to make a prediction, so improving interpretability. Benchmarking studies are conducted to illustrate competitiveness with the prediction accuracy of current state-of-the-art evolutionary ensemble learning algorithms, while providing solutions that are orders of magnitude simpler. Further benchmarking with a high cardinality dataset indicates that the proposed method is also more accurate and efficient than XGBoost.
\keywords{Boosting \and Stacking \and Genetic Programming.}
\end{abstract}
\section{Introduction}
Ensemble learning represents a widely employed meta-learning scheme for deploying multiple models under supervised learning tasks \cite{dietterich00,sipper22}. In general, there are three basic formulations: Bagging, Boosting or Stacking. We note, however, that most instances of ensemble learning adopt one scheme alone (\S \ref{sec:related}). Moreover, Boosting and Bagging represent the most widely adopted approaches.

In this work we investigate the utility of Boosting for constructing a Stacked ensemble. Thus, rather than members of an ensemble being deployed in parallel, our Stack assumes that ensemble members are deployed in the order that they were originally discovered. We therefore partner Stacking with a Boosting methodology for constructing the ensemble. This will enable us to incrementally develop ensemble classifiers that are explicitly focused on what previous members of the ensemble cannot do. In addition, we are then able to explicitly scale ensemble learning to large cardinality datasets. 

A benchmarking study demonstrates the effectiveness of the approach by first comparing with recent evolutionary ensemble learning algorithms on a suite of previously benchmarked low cardinality datasets. Having established the competitiveness of the proposed approach from the perspective of classifier accuracy, we then benchmark using a classification task consisting of 800 thousand records. The best of previous approaches is demonstrated not to scale under these conditions. Conversely, the efficiency of the proposed approach implies that parameter tuning is still feasible under the high cardinality setting.

The balance of the paper is organized as follows. Section \ref{sec:related} reviews key concepts from ensemble learning and summarizes research in evolutionary ensemble learning. The proposed approach for using Boosting to develop a Stacked ensemble is then detailed (\S \ref{sec:boost_overview}). The approach to benchmarking and parameterizing algorithms is then outlined (\S \ref{sec:benchmark}) and the results of two benchmarking studies detailed (\S \ref{sec:results}). Conclusions and future research completes the work (\S \ref{sec:conc}).

\section{Related Work}\label{sec:related}
Ensemble learning appears in three basic forms, summarized as follows and assuming (without loss of generality) that the underlying goal is to produce solutions for a classification problem: 

\begin{itemize}
\item \textbf{Bagging:} \emph{n} classifiers are independently constructed and an averaging scheme is adopted to aggregate the \emph{n} independent predictions into a single ensemble recommendation.\footnote{Classification tasks often assume the majority vote, although voting/weighting schemes might be evolved \cite{brameier01}.} The independence of the \emph{n} classifiers is established by building each classifier on a \emph{different} sample taken from the original training partition, or a `bag'. 
\item \textbf{Boosting:} constructs \emph{n} classifiers sequentially. Classifier performance is used to incrementally reweight the probability of sampling training data, $\mathcal{D}$, to appear in the training sample used to train the next classifier. Thus, each classifier encounters a sample of $\mathcal{D} / n$ training records. Post-training the ensemble again assumes an averaging scheme to aggregate the \emph{n} labels into a single label.
\item \textbf{Stacking:} assumes that a heterogeneous set of \emph{n} classifiers is trained on a common training partition. The predictions from the \emph{n} classifiers are then `stacked' to define a $n \times |\mathcal{D}|$ dataset that trains a `meta classifier' to produce the overall classification prediction.
\end{itemize}

Such a division of ensemble learning architectures reflects the view that the learning algorithm only constructs one model at a time. However, genetic programming (GP) develops multiple models simultaneously. Thus, one approach for GP ensemble learning might be to divide the population into islands and expose each island to different data subsets, where the data subset is constructed using a process synonymous with Bagging or Boosting (e.g. \cite{iba99,imamura03,folino08}). Some of the outcomes resulting from this research theme were that depending on the degree of isolation between populations, classifiers might result that were collectively strong, but individually weak \cite{imamura03}, or that solution simplicity might be enabled by the use of ensemble methods \cite{iba99}. Indeed, solution simplicity has been empirically reported for other GP ensemble formulations \cite{lichodzijewski10}.

Another recurring theme is to assume that an individual takes the form of a `multi-tree' \cite{soule99,muni04,badran12}. In this case, an `individual' is a team of \emph{n} Tree-structured programs. Constraints are then enforced on the operation of variation operators in order to maintain context between programs in the multi-tree. This concept was developed further under the guise of `levels of selection' in which selection can operate at the `level' of a program or ensemble \cite{thomason07,wu11}. However, in order to do so, it was necessary to have different performance definitions for each level. 

Virgolin revisits the theme of evolving ensembles under Bagging while concentrating on the role of selection \cite{virgolin21}. Specifically, individuals are evaluated w.r.t. \emph{all} bags. The cost of fitness evaluation over all bags is minimized by evaluating programs once and caching the corresponding fitness values. The motivation for evaluating performance over all the bags is to promote uniform development toward the performance goal.

Boosting and Bagging in particular have also motivated the development of mechanisms for decoupling GP from the cardinality of the training partition. In essence, GP performance evaluation is only ever conducted relative to the content of a data subset, $DS$, where $|DS| << |\mathcal{D}|$. However, the data subset is periodically resampled, with biases introduced to reflect the difficulty of labelling records correctly and the frequency of selecting data to appear in the data subset \cite{gathercole94,song05}. This relationship was then later explicitly formulated from the perspective of competitively coevolving ensembles against the data subset \cite{lichodzijewski08,lichodzijewski10,mcintyre11}, i.e. data subset and ensemble experience different performance functions.

Several coevolutionary approaches have also been proposed in which programs are defined in one population and ensembles are defined by another, e.g. \cite{lichodzijewski08,lichodzijewski10,mcintyre11,rodrigues20}. Programs are free to appear in multiple ensembles and the size of the ensemble is determined through evolution. 

The concept of `Stacking' has been less widely employed. However, the cascade-correlation approach to evolving neural networks might be viewed through this perspective \cite{fahlman89}. Specifically, cascade-correlation begins with a single `perceptron' and identifies the error residual. Assuming that the performance goal has not been reached, a new perceptron is added that receives as input all the previous perceptron outputs and the original data attributes. Each additional perceptron is trained to minimize the corresponding residual error. Potter and deJong demonstrated a coevolutionary approach to evolving neural networks using this architecture \cite{potter00}. Curry et al. benchmarked such an approach for training layers of GP classifiers under the cascade-correlation architecture \cite{curry07}. However, they discovered that the GP classifiers could frequently degenerate to doing no more than copy the input from the previous layer.

Finally, the concept of interpreting the output of a GP program as a dimension in a new feature space rather than a label is also of significance to several ensemble methods. Thus, programs comprising an ensemble might be rewarded for mapping to a (lower-dimensional) feature space that can be clustered into class-consistent regions \cite{mcintyre11,munoz15}. Multiple programs appearing in an ensemble define the dimensions of the new feature space. Likewise, ensembles based on multi-trees have been rewarded for mapping to a feature space that maximizes the performance of a linear discriminant classifier \cite{badran12}.

\section{Evolving an Ensemble Stack using Boosting}
The motivation of this work is to use a combination of boosting and stacking to address data cardinality while constructing the GP ensemble. Our insight is that classifiers can be sequentially added to the ensemble. After adding a classifier, the data correctly classified is removed from the training partition. Thus, as classifiers are added to the ensemble the cardinality of the training partition decreases. Moreover, the next classifier evolved is explicitly directed to label what the ensemble cannot currently label. Post training, the ensemble is deployed as a `stack' in which classifiers are deployed in the order in which they were evolved. As will become apparent, we need not visit all members of the ensemble stack in order to make a prediction. In the following, we first present the evolutionary cycle adopted for developing the Boosted Ensemble Stack (\S \ref{sec:boost_overview}) and then discuss how the Ensemble Stack is deployed (\S \ref{sec:deploy_stack}).

\subsection{The Boosting Ensemble Stack Algorithm}\label{sec:boost_overview}
Algorithm \ref{alg:cap} summarizes the overall approach taken to boosting in this research. The training dataset, $\mathcal{D}$, is defined in terms of a matrix $X^t$ of \emph{n} (input) records (each comprised of \emph{d}-attributes) and a vector $Y^t$ of \emph{n} labels (i.e. supervised learning for classification). We may sample records pairwise, $\langle \vec{x}_p \in X^t, y_p \in Y^t\rangle$, during training. The outer loop (Step \ref{alg:outer}) defines the number of `boosting epochs', where this sets an upper limit on ensemble size. Step \ref{alg:create_pop} initializes a new population, consisting of program decision stumps alone (single node programs). Step \ref{alg:evolve_progs} performs fitness evaluation, ranking, parent pool selection, and variation for a limited number of generations ({\ttfamily Max\_GP\_Epoch}). Specifically, the following form is assumed:

\begin{enumerate}
\item Fitness evaluation (Step \ref{alg:fitness}) assumes the Gini index in which the output of each program is modelled as a distribution (Algorithm \ref{alg:gini_fit}). Fitter individuals are those with a higher Gini index.
\item Parent pool selection (\emph{PP}) implies that the worst \%\emph{Gap} programs are deleted, leaving the parent pool as the survivors (Algorithm \ref{alg:cap}, Step \ref{alg:survive}). 
\item Test for early stopping (Step \ref{alg:exit_early}) where this is defined in terms of fitness and `bin purity' a concept defined below.
\item Variation operators (Step \ref{alg:offspring}) are limited to 1) cloning \%\emph{Gap} parents, 2) adding a single new node to a clone where the parameters of the new node are chosen stochastically, and 3) mutating any of the parameters in the resulting offspring.
\end{enumerate}

\begin{algorithm}
\caption{StackBoost($\langle X^t,Y^t \rangle$, {\ttfamily New\_Pop\_size, Max\_Boost\_epoch, Max\_GP\_epoch}). \emph{PP} is the parent pool}\label{alg:cap}
\begin{algorithmic}[1]

\State $Best\_Fit \gets 0$
\State $Ensemble \gets [\empty]$    \Comment{Initialize ensemble stack to null}

\For{$i \gets 1$ to {\ttfamily Max\_Boost\_epoch}}\label{alg:outer}
    \newline
    \State $best \gets$ {\ttfamily False}
    \State \emph{Pop} $\gets$ \emph{Initialize}({\ttfamily New\_Pop\_size})\label{alg:create_pop}	\Comment{Initialize a new pop}
    \newline
    \For{($j \gets 1$ to {\ttfamily Max\_GP\_epoch}) $\&$ ($best =$ {\ttfamily False})}\label{alg:evolve_progs}	\Comment{Evolve population}
        \State $Fitness \gets $[ $GiniIndexFitness(N,X^t,Y^t)$ for $N \in Pop$] \label{alg:fitness}
        \State $Ranked\_Pop \gets [\arg \mbox{ sort}(Fitness, Pop)]$
        \State $PP \gets Ranked\_Pop(Gap)$ \label{alg:survive}
        \newline
        \While{$Tree \in PP$}\label{alg:exit_early}									\Comment{Test for champion}
        		\State $Histogram, Interval \gets fitHist(Tree,X^t,Y^t)$\label{alg:histo}
	        	\If{($\exists Interval \in Histogram =$ Pure)}\label{alg:pure_exit}
			\If{($Tree.Fitness > Best\_Fit$)}\label{alg:inc_fit}
				\State $best \gets$ {\ttfamily True}							\Comment{Exit early if champion found}
				\State $Best\_Fit \gets Best\_Fit$
                \State $Champion\_Histogram \gets Histogram$
				\State $Champion \gets Tree$\label{alg:champ} 				\Comment{Record champion}
			\EndIf
		\EndIf
        \EndWhile
        \State \emph{Offspring} $\gets$ Variation($PP$, $\%Gap$) \label{alg:offspring}
        \State $Trees \gets PP\ \cup$ \emph{Offspring}
    \EndFor
    \newline
    
    \State $\langle X', Y' \rangle \gets \emptyset$
    \For {$Interval \in Champion\_Histogram$} \label{alg:mark}	\Comment{Identify Pure bin content}
         \If {$Interval = Pure$} 
                \State $\langle X', Y' \rangle \gets$ copy($\langle x_p, y_p \rangle \in Interval$)\label{alg:id_residual} \Comment{Identify correctly labelled data}
	\EndIf
    \EndFor
    \newline
    \State $\langle X^t, Y^t \rangle \gets \langle X^t, Y^t \rangle / \langle X', Y' \rangle$\label{alg:residual} \Comment{Define residual dataset}
    \State $Ensemble.push(Champion)$ \label{alg:append}		\Comment{Add \emph{Champion} to ensemble stack}
    \If {$\langle X^t, Y^t \rangle = \emptyset $}\label{alg:empty} \Comment{Case of early stopping}
       \State \Return $Ensemble$
    \EndIf
\EndFor
\State \Return $Ensemble$
\end{algorithmic}
\end{algorithm}

\begin{algorithm}
\caption{Gini Index Fitness ($Tree, X, Y$) returns gini index weighted by model complexity.}\label{alg:gini_fit}
\begin{algorithmic}[1]
\Require $Total(i) \gets$ \# records mapped to interval `\emph{i}'
\Require $Count(i,c) \gets$ \# class `\emph{c}' mapped to interval `\emph{i}'
\Require $\#Inst(c) \gets$ \# records of class `\emph{c}' in $Y$
\Require \emph{\%Used\_bins} is the \% of bins with a non-zero count
\For {$i \gets 1$ to {\ttfamily NumBin}}
	\For {$c \gets 1$ to {\ttfamily NumClass}}
		\If {$\mbox{Total}(i) \neq 0$}
			\State $hist(i, c) \gets \frac{Count(i, c)}{Total(i) \times \#Inst(c)}$
		\Else
			\State $hist(i, c) \gets 0$
		\EndIf
		\State $GiniIndex \gets GiniIndex + hist(i, c)^2 \times \#Inst(c)$ 
	\EndFor
\EndFor
\State \Return $GiniIndex + \alpha (\%Used\_bins)$
\end{algorithmic}
\end{algorithm}

\begin{figure}
\begin{center}
\includegraphics[width=10.5cm]{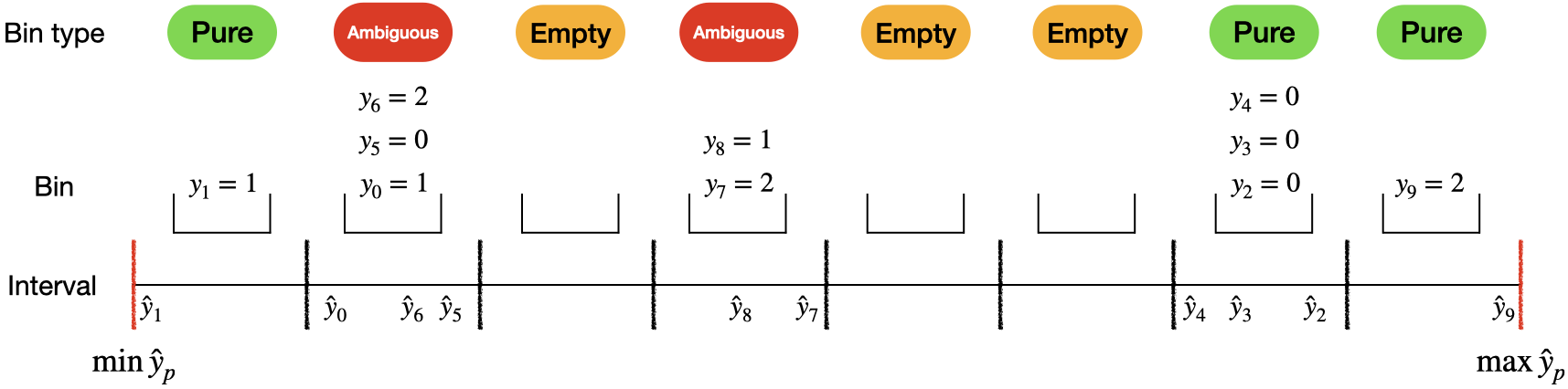}
\caption{Illustration for the relationship between program outputs ($\hat{y}_p$), intervals, bins, labels ($y_p = c$) and bin type.}\label{fig:distr}
\end{center}
\end{figure}

In assuming a performance function based on the Gini index, programs perform the following mapping: $\hat{y}_p = f(x_p)$ where $x_p$ is input record \emph{p} and $\hat{y}_p$ is the corresponding output from the program. There is no attempt to treat $\hat{y}_p$ as a predicted classification label for record \emph{p}. Instead, $\hat{y}_p$ represents the mapping of the original input, $x_p$, to a scalar value on a 1-dimensional number line, $\hat{y}$. After mapping all \emph{p} inputs to their corresponding $\hat{y}_p$ we quantize $\hat{y}$ into {\ttfamily NumBin} intervals (Step \ref{alg:histo}), as illustrated by Figure \ref{fig:distr}. Each interval defines an equal non-overlapping region of the number line $\hat{y}$ and an associated bin. The bins act as a container for the labels, $y_p = c$, associated with each $\hat{y}$ appearing in the bin's interval. Three types of the bin may now appear,

\begin{itemize}
\item \textbf{Empty bins:} have no $\hat{y}_p$ appearing in their interval.
\item \textbf{Pure bins:} have $\hat{y}_p$ appearing in their interval such that the majority of labels $y_p$ are the same. A `pure bin' assumes the label $y_p$ that reflects the majority of the bin content and is declared when the following condition holds,

\begin{equation}\label{eqn:purebin}
\frac{C_{bin} - y^*}{C_{bin}} < \beta 
\end{equation}

where $C_{bin}$ is the count of the number of records appearing in the bin, $C(c)$ represents the number of records of each class in the bin, and $y^* = \max_c C(c)$.
\item \textbf{Ambiguous bins:} imply that some $\hat{y}_p$ appear at an interval such that bin purity does not hold.
\end{itemize}

If a pure bin is encountered for an individual during training (Step \ref{alg:pure_exit}) that also has best fitness, then the next champion for including in the ensemble has been identified. Moreover, each time a new ensemble member is to be evolved, they have to improve on the fitness of the previous ensemble member (Step \ref{alg:inc_fit}).

Note that multiple Pure bins can appear, where this does not place any constraint on the labels representing different Pure bins. All training records corresponding to the `Pure bins' are then identified (Step \ref{alg:id_residual}) and removed to define a candidate residual dataset (Step \ref{alg:residual}).

Finally, the champion individual is then `pushed' to the list of ensemble members (Step \ref{alg:append}). Note that the order in which champions are pushed to the list is significant (\S \ref{sec:deploy_stack}). At the next iteration of the algorithm, an entirely new population of programs is evolved to specifically label what the members of the ensemble currently cannot label. Moreover, `early stopping' might appear if the residual dataset is empty (Step \ref{alg:empty}).

\subsection{Evaluating an Ensemble Stack Post Training}\label{sec:deploy_stack}

Post-training, we need to define how the ensemble operates collectively to produce a single label; whereas, during training, programs are constructed `independently', without direct reference to other members of the ensemble. Algorithm \ref{alg:stack_eval} summarizes how an ensemble is deployed as a `stack' and evaluated following the order in which each ensemble member was originally added to the ensemble, i.e. the stack chooses programs under a first-in, first-out order (Step \ref{alg:fifo_order}). Program \emph{i} may only suggest a label for bins that were originally identified as `Pure bins'. If the program produces a $\hat{y}_p$ corresponding to any other type of bin (Empty or Ambiguous), then the next program, $i + 1$, is called on to suggest its mapping $\hat{y}_p$ for record \emph{p} (Step \ref{alg:next_prog}). At any point where a program maps the input to a `Pure bin', then the corresponding label for that bin can be looked up. If this matches the actual label, $y_p$, then the ensemble prediction is correct (Step \ref{alg:correct}). This also implies that the entire ensemble need not be evaluated in order to make a prediction.

The first-in, first-out deployment of programs mimics the behaviour of `falling rule lists', a model of deployment significant to interpretable machine learning \cite{rudin19}. Thus, the most frequent queries are answered by the classifier deployed earliest. As the classes of queries become less frequent (more atypical) prediction is devolved to classifiers appearing deeper in the list (queue). The proposed evolutionary ensemble learner adds programs to the stack in precisely this order.

\begin{algorithm}
\caption{StackEvaluation($X,Y$, {\ttfamily Ensemble, stack\_depth})}\label{alg:stack_eval}
\begin{algorithmic}[1]
\State $\#correct \gets \#error \gets 0$
\For {$\langle x_p, y_p \rangle \in X, Y$}	\Comment{Loop over all data partition}
   \State \emph{Initialize.Stack}({\ttfamily Ensemble})\label{alg:fifo_order} \Comment{Initialize stack, using original `fifo' order}
   \State $n = $ {\ttfamily stack\_depth}	\Comment{Initialize to max ensemble members}
   \Repeat
	\State \emph{Tree $\gets$ pop}(\emph{Stack})	\Comment{Pop a tree, oldest first}
	\State $\langle bin_p, \hat{y}_p \rangle \gets Tree(x_p)$	\Comment{Execute Tree on record $x_p$}
	\If {$bin_p$ is Pure}			\Comment{Test for a `Pure' bin type}
	   \State $n \gets 0$			\Comment{Pure bin, so use this Tree for prediction}
	   \If {$\hat{y}_p \neq y_p$}	\Comment{Case of prediction not matching class}
	      \State $\#error++$		\Comment{Update classification performance}
	   \Else
	      \State $\#correct++$\label{alg:correct}
	   \EndIf
	\Else
	   \State $n--$\label{alg:next_prog}		\Comment{Not a pure bin, so next Tree}
	\EndIf
   \Until {$n$ is 0}				\Comment{No further trees in Ensemble}
\EndFor
\State \Return $\langle \#error, \#correct \rangle$
\end{algorithmic}
\end{algorithm}

\subsection{Using an Extremely Large Number of Bins}

One approach to parameterizing BStacGP is to force the number of bins to be very low (2 or 3). The intuition of is that this rewards BStacGP discovering a mapping of records to bins that develops at least one bin to be pure. This approach will be adopted later for the ‘small scale’ benchmark. Under high cardinality datasets the opposite approach is assumed. The insight behind this is that there is sufficient data to support multiple pure bins such that we maximize the number of training records mapped to pure bins by the same program. This will also result in the fastest decrease in data cardinality during training. 

Taking this latter approach to the limit, we assume the number of bins is set by the size of a floating point number or $2^{32}$. During training, the training data partition again defines bins as pure, ambiguous or empty. However, given the resolution of the bins, most bins should actually be `pure’, reducing the number of boosting epochs necessary to build the ensemble. Under test conditions, given that the bins have a high resolution, it is also likely that test data will be mapped to ‘empty’ bins. Hence, under test conditions, the test data is labelled by the pure bin that it is closest to. If the closest bin is not pure but ambiguous, then the next tree in the stack is queried.



\section{Experimental Methodology}\label{sec:benchmark}
A benchmarking comparison will be made against a set of five datasets from a recent previous study \cite{virgolin21}. This enables us to establish to what degree the proposed approach is competitive with five state-of-the-art approaches for GP ensemble classification. The datasets appearing in this comparison are all widely used binary classification tasks from the UCI repository,\footnote{url{https://archive-beta.ics.uci.edu}} as summarized by Table \ref{tbl:datasets}. The training and test partitions are stratified to provide the same class distribution in each partition and the training/test split is always 70/30\%.

The algorithms appearing in this comparison take the following form:

\begin{itemize}
\item \textbf{2SEGP:} A bagging approach to evolving GP ensembles in which the underlying design goal is to maintain uniform performance across the multiple bootstrap bags. Such an approach was demonstrated to be particularly competitive with other recent developments in evolutionary ensemble learning \cite{virgolin21}. In addition, the availability of a code base enables us to make additional benchmarking comparisons under a high cardinality dataset.
\item \textbf{eGPw:} Represents the best-performing configuration of the cooperative coevolutionary approach to ensemble learning proposed in \cite{rodrigues20}. Specifically, benchmarking revealed the ability to discover simple solutions to binary classification problems while also being competitive with Random Forests and XGBoost.
\item \textbf{DNGP:} Represents an approach to GP ensembles in which diversity maintenance represents the underlying design goal \cite{wang19}. Diversity maintenance represents a reoccurring theme in evolutionary ensemble learning, where the underlying motivation is to reduce the correlation between ensemble members. The framework was reimplemented and deployed in the benchmarking study of \cite{virgolin21}.
\item \textbf{M3GP:} Is not explicitly an evolutionary ensemble learner but does evolve a set of programs to perform feature engineering \cite{munoz15}. The underlying objective is to discover a mapping to a new feature space that enables clustering to separate between classes. M3GP has been extensively benchmarked in multiple contexts and included in this study as an example of what GP classification can achieve without evolutionary ensemble learning being the design goal \cite{munoz15,cava19}.
\end{itemize}

\begin{table}
\caption{Properties of the benchmarking datasets. Class distribution reflects the distribution of positive to negative class instances.}\label{tbl:datasets}
\begin{center}
\begin{tabular}{|c|c|c|c|c|}\hline

Dataset & \# Features & Cardinality & Class distribution \\
 & (\emph{d}) & ($|\mathcal{D}|$) & (\%P / \%N) \\ \hline
 \multicolumn{4}{|c|}{Small benchmarking datasets} \\ \hline
BCW & 11 & 683 & 30/65 \\
HEART & 13 & 270 & 45/55 \\
IONO & 33 & 351 & 65/35 \\
PARKS & 23 & 195 & 75/25 \\
SONAR & 61 & 208 & 46/54 \\ \hline
 \multicolumn{4}{|c|}{Large benchmarking dataset} \\ \hline
CTU &8 &801132 &55/45 \\ \hline
\end{tabular}
\end{center}
\label{default}
\end{table}%

The second benchmarking study is then performed on a large cardinality dataset describing an intrusion detection task. This dataset is the union of the normal and botnet data from the CTU-13 dataset \cite{garcia14}, resulting in hundreds of thousands of data records (Table \ref{tbl:datasets}). In this case, we compare the best two evolutionary ensembles from the first study with C4.5 \cite{quinlan93} and XGBoost \cite{chen16}. The latter represent very efficient non-evolutionary machine learning approaches to classification.

\begin{table}
    \centering
    \caption{BStacGP parameterization. \emph{Gap} defines the size of the parent pool. $\beta$ is defined by Eqn. (\ref{eqn:purebin}). $\alpha$ weights the fitness regularization term, Algorithm \ref{alg:gini_fit}. {\ttfamily NumBin} appears in Algorithm \ref{alg:stack_eval} and the remaining parameters in Algorithm \ref{alg:cap}}
    \label{tab:bstac_param}
    \begin{tabular}{|c|c|c|c|c|} \hline
        Benchmarking Study & \multicolumn{2}{|c|}{ Small Scale } & \multicolumn{2}{|c|}{ Large Scale } \\ \hline
        Parameter & fast & slow & fast & slow \\ \hline
        {\ttfamily Max\_Boost\_epoch} &  \multicolumn{2}{|c|}{ 1000 } &  \multicolumn{2}{|c|}{ 10 } \\ \hline
        {\ttfamily Max\_GP\_epoch} & \multicolumn{2}{|c|}{ 30 } & 3 & 6\\ \hline
        {\ttfamily New\_Pop\_Size} & 30 & 1000 & \multicolumn{2}{|c|}{ 30 } \\ \hline
        \emph{Gap} & 10 & 300 & \multicolumn{2}{|c|}{ 10 }\\ \hline
        {\ttfamily NumBin} & \multicolumn{2}{|c|}{ 2 } & \multicolumn{2}{|c|}{ $2^{32}$ }\\ \hline
        Bin Purity ($\beta$) & \multicolumn{2}{|c|}{ 0.99 } & 0.6 & 0.75 \\ \hline
        Regularization ($\alpha$) & \multicolumn{2}{|c|}{ 0.0 } & \multicolumn{2}{|c|}{ 0.4 } \\ \hline
        Num. Trials & \multicolumn{4}{|c|}{ 40 } \\ \hline
    \end{tabular}
\end{table}

Table \ref{tab:bstac_param} summarizes the parameters assumed for BStacGP. Two parameterizations are considered in each benchmarking study: `slow and complex' versus `fast and simple'. One insight is that higher data cardinality can imply a higher bin count. We therefore use the lowest bin count (2) and a high purity threshold (0.99) as a starting point under the Small Scale benchmarking study. Such a combination forces at least one of the 2 bins to satisfy the high bin purity threshold. We then differentiate between `slow and complex' versus `fast and simple' scenarios by increasing the population size (with the parent pool increasing proportionally). The value for {\ttfamily Max\_Boost\_epoch} is set intentionally high, where in practice such an ensemble size is not encountered due to early stopping being triggered (Algorithm \ref{alg:cap}, Step \ref{alg:empty}). Under the large scale benchmark, the largest bin count was assumed, where this does not imply that this number of bins needs to contain values, but it does mean that the distribution has the most resolution.

\section{Results}\label{sec:results}
Two benchmarking studies are performed.\footnote{Laptop with Intel i7 10700k CPU, 4.3GHz single core.} The first assumes a suite of `small scale' classification tasks (\S \ref{sec:small}) that recently provided the basis for comparing several state-of-the-art GP evolutionary ensemble learners \cite{virgolin21}. The second study reports results on a single large cardinality dataset using the best two evolutionary ensemble learners from the first study, and two non-evolutionary methods (\S \ref{sec:big}). Hereafter, the proposed approach is referred to as BStacGP.

\subsection{Small Scale Classification Tasks}\label{sec:small}

Tables \ref{tab:small_datasets_train} and \ref{tab:small_datasets_test} report benchmarking for the five small datasets. In the previous benchmarking study, 2SEGP and M3GP were the best-performing algorithms on these datasets \cite{virgolin21}. Introducing the proposed Boosted Stack ensemble (BStacGP) to the comparison changes the ranking somewhat. That said, all five GP formulations perform well on the BCW dataset ($>95\%$ under test), whereas the widest variance in classifier performance appears under HEART and SONAR. Applying the Friedman non-parametric test for multiple models to the ranking of test performance fails to reject the null hypothesis (all algorithms are ranked equally). Given that these are all strong classification algorithms this is not in itself unexpected.

\begin{table}[h]
\caption{Classifier accuracy on the \emph{training partition} for small binary datasets. Bold indicates the best-performing classifier on the dataset}
\label{tab:small_datasets_train}
\centering
\begin{tabular}{|c|c|c|c|c|c|c|} \hline
Dataset & \multicolumn{2}{|c|}{BStacGP} & 2SEGP & DNGP & eGPw & M3GP \\
& slow & fast & \cite{virgolin21} & \cite{virgolin21} & \cite{virgolin21} & \cite{virgolin21} \\ \hline
BCW & 0.994 & \textbf{0.995} & \textbf{0.995} & 0.979 & 0.983 & 0.971 \\
 & $\pm$ 0.006 & $\pm$ 0.005 & $\pm$ 0.005 & $\pm$ 0.010 & $\pm$ 0.008 & $\pm$ 0.002 \\ \hline
HEART & \textbf{1.000} & 0.985 & 0.944 & 0.915 & 0.907 & 0.970 \\
 & $\pm$ 0.0 & $\pm$ 0.015 & $\pm$ 0.022 & $\pm$ 0.021 & $\pm$ 0.025 & $\pm$ 0.017 \\ \hline
IONO & \textbf{0.993} & 0.983 & 0.976 & 0.955 & 0.884 & 0.932 \\
 & $\pm$ 0.008 & $\pm$ 0.017 & $\pm$ 0.017 & $\pm$ 0.015 & $\pm$ 0.032 & $\pm$ 0.042 \\ \hline
PARKS & 0.982 & \textbf{0.996} & 0.948 & 0.931 & 0.923 & 0.981 \\
 & $\pm$ 0.018 & $\pm$ 0.004 & $\pm$ 0.011 & $\pm$ 0.057 & $\pm$ 0.042 & $\pm$ 0.024 \\ \hline
SONAR & 0.999 & \textbf{1.000} & 0.966 & 0.924 & 0.924 & \textbf{1.000} \\ 
& $\pm$ 0.001 & $\pm$ 0.0 & $\pm$ 0.034 & $\pm$ 0.043 & $\pm$ 0.034 & $\pm$ 0.012 \\ \hline
\end{tabular}
\end{table}

\begin{table}
\caption{Classifier accuracy on the \emph{test partition} for small binary datasets. Bold indicates the best-performing result}
\label{tab:small_datasets_test}
\centering
\begin{tabular}{|c|c|c|c|c|c|c|} \hline
Dataset & \multicolumn{2}{|c|}{BStacGP} & 2SEGP & DNGP & eGPw & M3GP \\
& slow & fast & \cite{virgolin21} & \cite{virgolin21} & \cite{virgolin21} & \cite{virgolin21} \\ \hline
BCW & \textbf{0.96} & 0.957 & \textbf{0.965} & 0.959 & 0.956 & 0.957 \\
 & $\pm$ 0.017 & $\pm$ 0.022 & $\pm$ 0.018 & $\pm$ 0.019 & $\pm$ 0.018 & $\pm$ 0.014 \\ \hline
HEART & 0.803 & 0.796 & \textbf{0.815} & \textbf{0.815} & 0.790 & 0.778 \\
 & $\pm$ 0.094 & $\pm$ 0.052 & $\pm$ 0.062 & $\pm$ 0.049 & $\pm$ 0.034 & $\pm$ 0.069 \\ \hline
IONO & \textbf{0.924} & 0.901 & 0.896 & 0.901 & 0.830 & 0.871 \\
 & $\pm$ 0.027 & $\pm$ 0.047 & $\pm$ 0.047 & $\pm$ 0.026 & $\pm$ 0.057 & $\pm$ 0.057 \\ \hline
PARKS & \textbf{0.951} & 0.937 & 0.936 & 0.917 & 0.822 & 0.897 \\
 & $\pm$ 0.017 & $\pm$ 0.013 & $\pm$ 0.012 & $\pm$ 0.055 & $\pm$ 0.064 & $\pm$ 0.051 \\ \hline
SONAR & 0.76 & 0.728 & 0.738 & 0.730 & 0.762 & \textbf{0.810} \\ 
 & $\pm$ 0.083 & $\pm$ 0.131 & $\pm$ 0.067 & $\pm$ 0.063 & $\pm$ 0.060 & $\pm$ 0.071 \\ \hline \hline
Av.Rank & 2.0 & 3.8 & 2.7 & 3.2 & 5.0 & 4.3 \\ \hline
\end{tabular}
\end{table}

\begin{figure}
\begin{center}
\includegraphics[width=9cm]{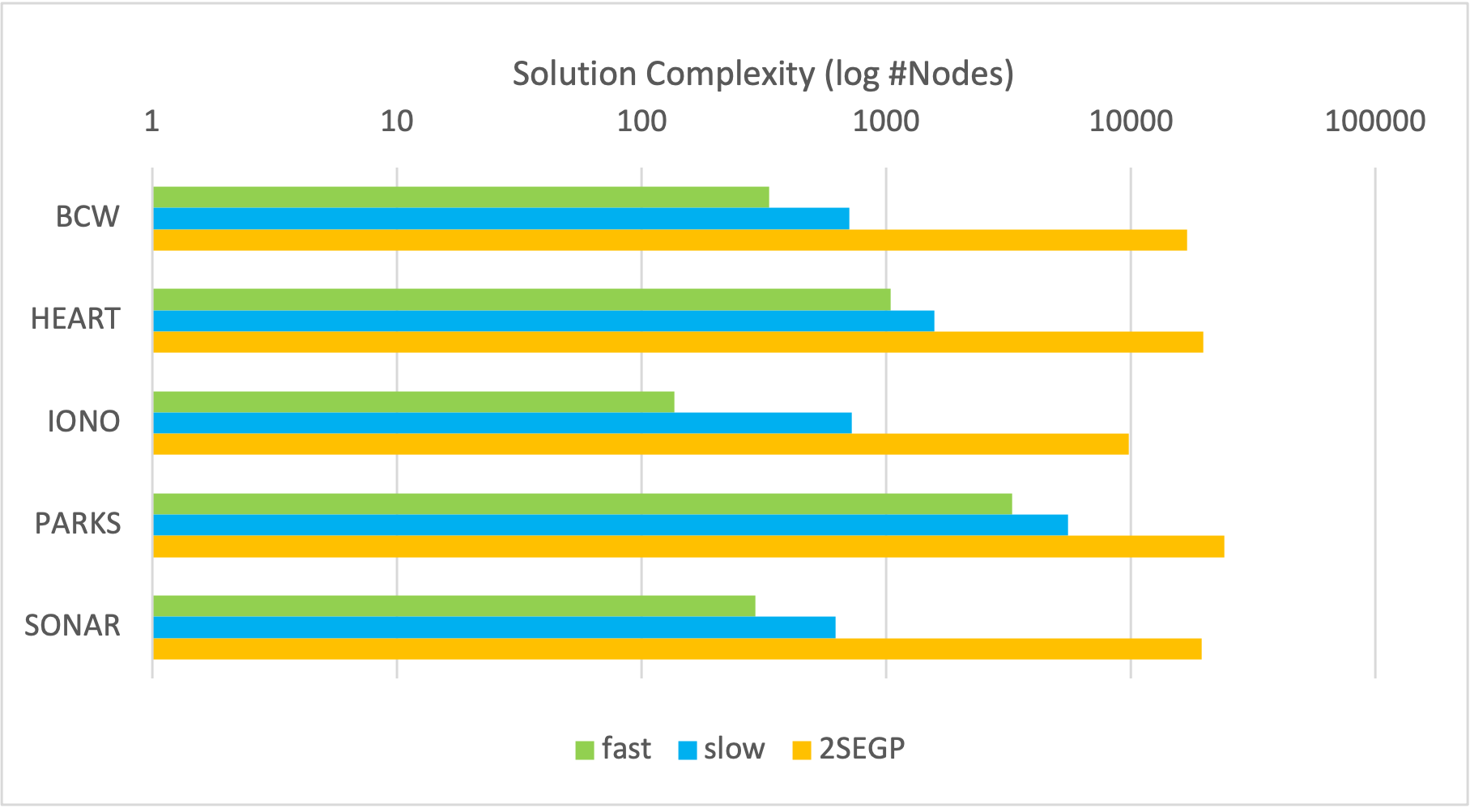}
\caption{Solution Complexity for BStacGP and 2SEGP on small scale classification tasks. `fast' and `slow' represent the two BStacGP parameterizations.}\label{fig:complex}
\end{center}
\end{figure}

\begin{figure}
\begin{center}
\includegraphics[width=9cm]{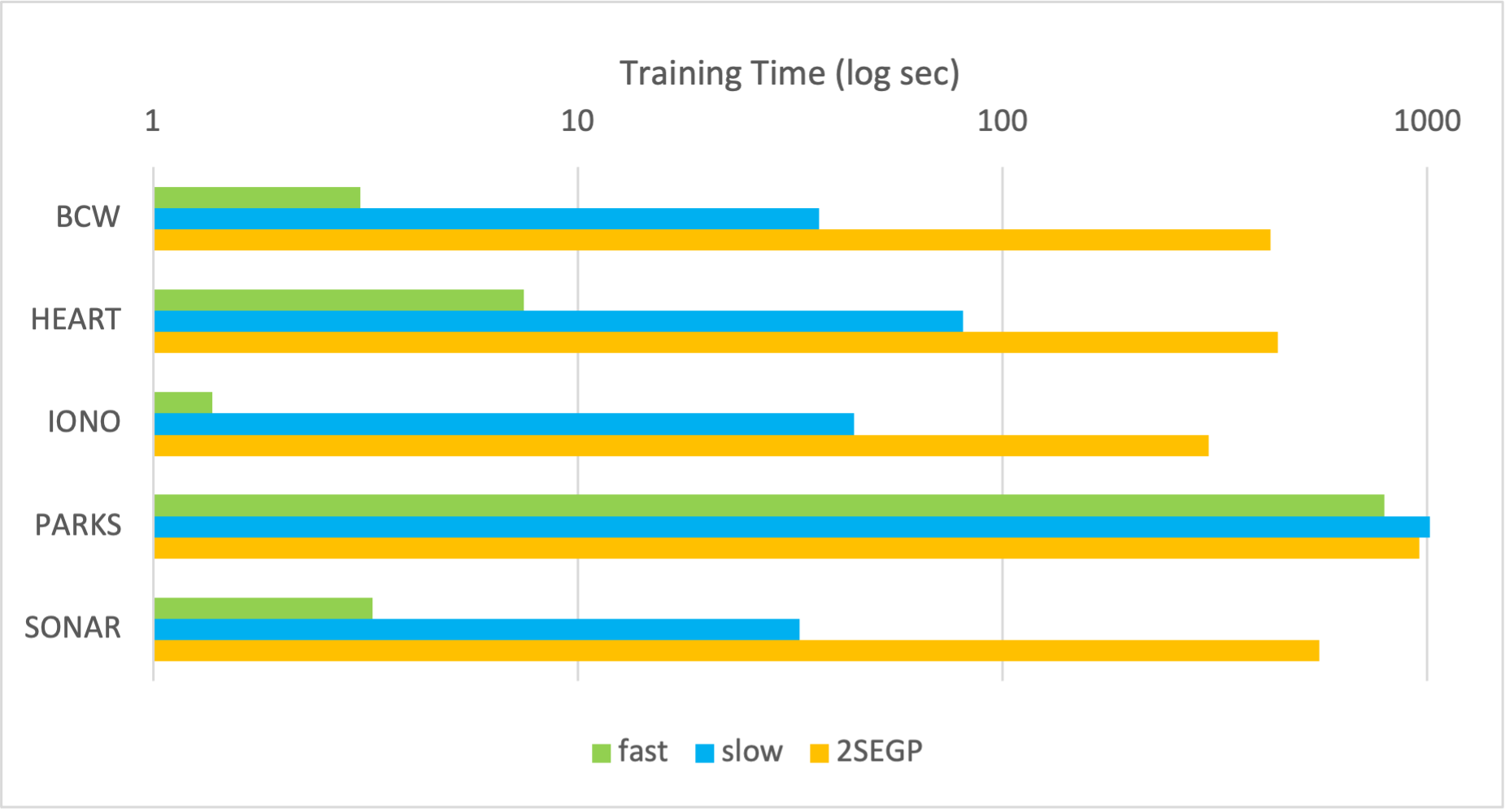}
\caption{Training time for BStacGP and 2SEGP on small scale classification tasks. `fast' and `slow' represent the two BStacGP parameterizations.}\label{fig:time}
\end{center}
\end{figure}

However, BStacGP and 2SEGP are the two highest ranked algorithms on training and test. With this in mind we consider the average solution complexity and time to evolve solutions. In the case of complexity, the average number of nodes in the Tree structured GP individuals comprising an ensemble is counted, Figure \ref{fig:complex}. It is apparent that BStacGP is able to discover solutions that are typically an order of magnitude simpler.  Figure \ref{fig:time} summarizes the wall clock time to conduct training. Both BStacGP and 2SEGP are implemented in Python. BStacGP typically completes training an order of magnitude earlier than 2SEGP. In summary, the process by which BstackGP incrementally only evolves classifiers against the `residual' misclassified data results in a significant decrease in computational cost and model complexity.

\subsection{Large Scale Classification Task}\label{sec:big}
BStacBP and 2SEGP accounted for 4 out of 5 of the best classifier performance under the test partition of the `Small Scale' task (Table \ref{tab:small_datasets_test}). With this in mind, we now compare the performance of BStacBP and 2SEGP under the high-cardinality CTU dataset with the non-evolutionary classifiers of C4.5 and XGBoost. Table \ref{tab:large_data_comparison} reports performance for the average (over 40 trials) of BStacBP, C4.5 and XGBoost and a single run of 2SEGP. Specifically, the high run time cost of 2SEGP precluded performing multiple trials on this benchmark.\footnote{2SEGP parameterization: pop. size 500, ensemble size 50, max. tree size 500.} The best-performing algorithm under test is XGBoost, however, BStacGP is within $1.3\%$ of this result. Conversely, 2SEGP returns a 10\% lower accuracy. This result is undoubtedly in part because it took 4 hours to perform 20 generations.\footnote{2SEGP used 100 generations in the small scale task.} 

In the case of solution complexity (as measured by node count), BStacGP returns solutions that are 2 to 3 orders of magnitude simpler than any comparator algorithm. From the perspective of the computational cost of performing training, C4.5 is the fastest. However, of the ensemble algorithms, BStacGP is twice as fast as XGBoost and 3 orders of magnitude faster than 2SEGP.

\begin{table}
\caption{CTU dataset comparison assuming the `slow' parameterization}
\label{tab:large_data_comparison}
\centering
\begin{tabular}{|l|l|l|l|l|}
\hline
Algorithm & BStacGP    & 2SEGP       & Decision Tree & XGBoost    \\ \hline
Train Accuracy & 0.987 & 0.8437 & \textbf{0.9986}        & 0.9716 \\ \hline
Test Accuracy  & 0.953 & 0.8419 & 0.9625        & \textbf{0.9681} \\ \hline
number of nodes      & \textbf{157.9}   & 8454      & 52801         & 26792  \\ \hline
number of trees & 2.87 & 50 & \textbf{1} & 300 \\ \hline
avg. tree depth & \textbf{6.03} & -- & 55 & \textbf{6} \\ \hline
time (sec)   & 23.09  & 13160.92    & \textbf{1.54}          & 46.92  \\ \hline
\end{tabular}
\end{table}

A second experiment is performed using the CTU dataset, the hypothesis, in this case, being that constraining C4.5 and XGBoost to the same complexity as BStacGP will have a significant impact on their classification performance. Put another way, the comparator algorithms will not be able to discover solutions with similar complexity to BStacGP without significantly compromising their classification accuracy.

\begin{table}
\caption{CTU dataset low complexity comparison. BStacGP assumes the fast parameterization from Table \ref{tab:bstac_param}}
\label{tab:large_data_low_complexity}
\centering
\begin{tabular}{|l|l|l|l|l|}
\hline
Algorithm & BStacGP & Decision Tree & XGBoost  \\ \hline
Train Accuracy & \textbf{0.985}                & 0.914                         & 0.8796 \\ \hline
Test Accuracy  & \textbf{0.952}                & 0.915                         & 0.8795 \\ \hline
number of nodes      & \textbf{47.55}                  & 59                             & 75     \\ \hline
number of trees & 2.25 & \textbf{1} & 5 \\ \hline
avg. tree depth & 3.8 & 8 & \textbf{3} \\ \hline
time (sec)           & 11.28                  & 0.83 & \textbf{0.41}   \\ \hline
\end{tabular}
\end{table}

Table \ref{tab:large_data_low_complexity} summarizes performance under the low complexity condition. It is now apparent that BStacGP is able to maintain solution accuracy on this task as well as further reduce the computation necessary to identify such a solution. This implies that BStacGP has the potential to scale to tasks that other formulations of evolutionary ensemble learning fail to scale to while maintaining state-of-the-art classification performance. Thus, as dataset cardinality increases algorithm efficiency has an increasing impact, as it is simply not possible to tune parameters, an important practical property to have.

Post training, BStacGP solutions can be queried. For example, the first program from a typical BStack stack ensemble provided $\approx 57\%$ of the labels. The second $\approx 24\%$, the third $\approx 15\%$ and the forth $\approx 4\%$. This illustrates the `fall through' nature of BStacGP operation in which most of the data is labeled by a single GP program. The complexity of trees associated with each stack level are respectively 138, 250, 305 and 407 nodes, again illustrating how complexity increases with position in the stack.

\section {Conclusion}\label{sec:conc}
An approach to evolutionary ensemble learning is proposed that employs boosting to develop a stack of GP programs. Key components of the framework include: 1) interpreting the program output as a distribution; 2) quantizing the distribution into intervals and therefore `binning' the number of records mapped to an interval; 3) making predictions on the basis of `bin purity'; and 4) removing records from the training partition corresponding to correctly classified instances. The combination of these properties incrementally reduces the cardinality of the training partition as classifiers are added to the ensemble, and explicitly focuses the role of the next program on what previous programs could not classify. Moreover, the resulting ensemble is then deployed as a `Stack'. This is important because it now means that only part of the ensemble is responsible for providing a label. Such an approach may improve the explainability of the resulting ensemble.

Accuracy of the BStacGP framework on small cardinality datasets previously employed for benchmarking is empirically shown to be comparable to state-of-the-art evolutionary ensemble learners. Moreover, training time and model simplicity is significantly improved. This property is shown to be key to scaling BStacGP much more efficiently to a large cardinality dataset containing hundreds of thousands of records. Indeed the results are competitive with non-evolutionary methods.

Future work will scale BStacGP to multi-class classification and continue to investigate scalability and solution transparency.
\subsubsection{Acknowledgements} Please place your acknowledgments at
the end of the paper, preceded by an unnumbered run-in heading (i.e.
3rd-level heading).

%
%
%
\bibliographystyle{splncs04}
\bibliography{evoStack}
\end{document}